\documentclass[11pt,a4paper]{article}
\PassOptionsToPackage{breaklinks}{hyperref}
\usepackage[hyphens]{xurl}
\usepackage[hyperref]{eacl2021}
\usepackage[hyphenbreaks]{breakurl}
\usepackage{times}
\usepackage{latexsym}

\usepackage{microtype}

\aclfinalcopy 

\usepackage{graphicx}
\usepackage{tabularx}
\usepackage{multirow}
\usepackage{comment}
\usepackage{booktabs}
\usepackage{soul}

\usepackage{setspace}

\usepackage{epstopdf}
\usepackage[latin1]{inputenc}
\usepackage{xstring}

\usepackage{csquotes}
\renewcommand{\mkbegdispquote}[2]{\itshape}

\newcolumntype{C}[1]{>{\centering\let\newline\\\arraybackslash\hspace{0pt}}m{#1}}
\newcolumntype{R}[1]{>{\raggedleft\let\newline\\\arraybackslash\hspace{0pt}}m{#1}}
\newcolumntype{L}[1]{>{\raggedright\let\newline\\\arraybackslash\hspace{0pt}}m{#1}}
\usepackage{relsize}

\title{WiC-TSV: An Evaluation Benchmark for \\ Target Sense Verification of Words in Context}

\author{Anna Breit\textsuperscript{1}, Artem Revenko\textsuperscript{1}, Kiamehr Rezaee\textsuperscript{2}, \\ \textbf{Mohammad Taher Pilehvar\textsuperscript{3}, Jose Camacho-Collados\textsuperscript{4}} \\
\textsuperscript{1}Semantic Web Company, Austria;
        \textsuperscript{2}Iran University of Science and Technology; \\ 
        \textsuperscript{3}Tehran Institute for Advanced Studies, Iran; 
        \textsuperscript{4}Cardiff University, United Kingdom \\ 
        \textsuperscript{1}\tt \{anna.breit,artem.revenko\}@semantic-web.com,
        \\ \textsuperscript{2}\tt k\_rezaee@comp.iust.ac.ir, \textsuperscript{3}\tt mp792@cam.ac.uk, \\ \textsuperscript{4}\tt camachocolladosj@cardiff.ac.uk}

\date{}

\begin{document}
\maketitle
\begin{abstract}
We present WiC-TSV, a new multi-domain evaluation benchmark for Word Sense Disambiguation. 
More specifically, we introduce a framework for \textit{Target Sense Verification of Words in Context} which grounds its uniqueness in the formulation as binary classification task thus being independent of external sense inventories, and the coverage of various domains.
This makes the dataset highly flexible for the evaluation of a diverse set of models and systems in and across domains. 
WiC-TSV provides three different evaluation settings, depending on the input signals provided to the model.
We set baseline performance on the dataset using state-of-the-art language models.
Experimental results show that even though these models can perform decently on the task, there remains a gap between machine and human performance, especially in out-of-domain settings. 
WiC-TSV data is available at  \url{https://competitions.codalab.org/competitions/23683}
\end{abstract}

\section{Introduction}
\label{introduction}

Word Sense Disambiguation (WSD) is a long-standing task in Natural Language Processing and Artificial Intelligence. While progress has been made in recent years, mainly thanks to the surge of transformer-based language models such as BERT \cite{loureiro-jorge-2019-language,vial2019sensecompresswsd,huang-etal-2019-glossbert}
, the evaluation of WSD models has been limited to a set of (mostly SemEval-based) standard WSD datasets \cite{raganato-camachocollados-navigli:2017:EACLlong}. 
These datasets usually come in one of the two forms: \textit{lexical sample}, in which a target word is placed in various contexts, triggering different senses, and \textit{all-words}, in which all the content words in a given text are to be disambiguated.
Both settings, however, come with a major restriction: word senses in the datasets are linked to external sense inventories such as WordNet \cite{Fellbaum:98}.
Therefore, existing benchmarks are limited to only those WSD systems in which sense distinctions are defined according to an underlying sense inventory.
This not only gives restrictions to the model's flexibility, 
but also enforces the assumption of the availability of complete data.
However, as general sense inventories are complex to maintain they often lag behind in being up-to-date\footnote{The last update in WordNet dates back to June 2011.}, yielding to the absence of novel terms and term usages. Furthermore, the coverage of domain-specific terms and named entities in general sense inventories is quite limited, while domain-specific sense inventories are rare and in most cases incomplete. 

As a motivating example, let us assume \textit{Technology} as the target domain and the collection of information on the current technology landscape as a goal. Therefore, the following context needs to be disambiguated in order to evaluate its relevance: 
\begin{displayquote}
From 1970 to 2007, \textbf{Apple's} chief executive was former Beatles road manager Neil Aspinall.
\end{displayquote}
Even when incorporating a general sense inventory (which would include senses for the fruit and the tree) and a technology-specific sense inventory (which would include the sense for Apple Inc. the technology company), the actual target sense of this context (i.e., Apple Corps Limited, a multimedia corporation founded by the Beatles) may still be missing, which makes the annotation of the correct sense impossible. For these reasons, the current WSD task formulation and existing benchmarks are not fully able to evaluate the suitability of disambiguation systems in realistic domain-specific and/or enterprise settings.


In this paper, we try to fill this gap by proposing a re-formulation of the existing WSD task as well as a new benchmark for evaluating WSD systems under this paradigm. 

\textbf{Target Sense Verification (TSV)} formulates the disambiguation of a word as a binary classification task where the equivalence of the intended sense of a word in context and a single given sense is evaluated. For instance, in the example above, the system would need to decide whether the sentence refers to \textit{Apple Inc.} the technology company or not, by being provided with a sense indicator for solely Apple Inc. (e.g., the hypernym \textit{technology company} or the definition).

A system able to efficiently solve the TSV task 
could be effectively used in the scenario of collecting and tagging large amounts of textual data; e.g., from social media, news agencies, blogs and for downstream tasks such as information retrieval, sentiment analysis or relation extraction. Furthermore, such a system could be a good candidate for entity linking (EL) as the task statement of TSV resembles the usage of enterprise knowledge graphs \cite{Galkin2017} for EL: typically, small domain-specific enterprise knowledge graphs only contain entities from the domain of interest, partially or completely missing the general purpose senses of the contained labels.

In order to train and evaluate models for TSV we constructed \textbf{WiC-TSV (Word in Context - Target Sense Verification)} a multi-domain dataset and evaluated standard unsupervised and supervised approaches (including language models). While WiC-TSV's training and development set consist of general purpose instances, the test set contains domain-specific instances from three different domains. Therefore, this dataset aims at evaluating the ability of a model to (1) disambiguate the word in context without an external sense inventory, (2)  deal with unseen instances and incomplete data, and (3) transfer the intrinsic knowledge (gained on general domain data) into a specific domain.

\section{Related Work}
\label{related_work}

\paragraph{Word Sense Disambiguation.} 
The task of WSD consists of associating a word in context with its most appropriate entry in a given sense inventory \cite{navigli:09}, e.g., WordNet. 
For WSD there are many associated datasets \cite{raganato-camachocollados-navigli:2017:EACLlong,vialetal:18,Roder2018,ling-etal-2015-design}, 
including domain-specific ones \cite{agirre-etal-2009-semeval,faralli-navigli-2012-new}. 
The main difference between WSD and its re-formulation TSV is that for TSV the availability of a sense inventory is not required. Instead of associating a word in context with its most appropriate sense, the usage of a single given sense in the provided context is to be verified. Systems that aim to solve the proposed task are therefore not required to model all senses of the target word, but only a single sense instead.

This facilitates the development of systems for specific domains or settings, as no general-domain knowledge resource is required to perform this task. 
For instance, an Indonesian company may want to retrieve all sentences referring to the \textit{Java} island and not other unrelated senses. This framing of the task is frequent in business and data mining settings where domain-specific knowledge resources or inventories may be available, without the need for modeling instances from other domains.



\paragraph{WiC.}
The task closest to the proposed WiC-TSV is probably Word-in-Context \cite[WiC]{pilehvar-camacho-collados-2019-wic}, which our dataset is based on. WiC is a binary classification dataset where a target word is presented within two different contexts. The task consists of deciding whether the word is associated with the same sense in the two contexts or not. 
WiC is also one of the tasks included in the general language understanding framework SuperGLUE \cite{wang2019superglue}. 

WiC-TSV inherits some of the desirable properties of WiC, such as independence from external sense inventories and the binary classification nature of the task.
However, though our benchmark draws ideas from the Word-in-Context benchmark, 
it provides a different evaluation setting with additional flavors. 
The main difference with respect to our dataset lies in the presence of relevant information such as hypernyms and definitions, which makes our dataset more realistic and a direct proxy for downstream evaluation: in WiC-TSV the ambiguous target word in a single context is compared against a specific target sense (indicated by provided hypernyms and definitions), in contrast to the comparison of the intended senses of the target word in two different contexts.
Also, the task is more targeted at word-level representation, as in one of the tasks (i.e. {\it hypernymy} task) the model is not provided with any contextual information and, therefore, needs to have a clear understanding of the word to be able to make correct judgements.
Moreover, WiC-TSV includes instances from three domains (cocktails, medicine, and computer science) in its test set, which makes the benchmark more challenging and comparable to a real setting.


\section{WiC-TSV: The Benchmark} 
\label{benchmark}


A goal of this benchmark is to evaluate the ability of a model to
verify the target sense of
a word in a context without the usage of an external sense inventory, i.e., without knowing all possible senses of the target word. Another model quality that is aimed at with the presented benchmark is the ability to transfer the intrinsic knowledge into a specific domain. As for most areas, domain-specific training data is hard to obtain, being able to learn on general purpose data and still perform well on domain-specific data is a huge advantage in a real world setting. 

To this end, we constructed a benchmark satisfying following requirements:
\begin{enumerate}
    \item Knowledge of only a \textbf{single sense} of the target word;\label{req1}
    \item Knowledge of the \textbf{definition and/or hypernyms} of the target sense; \label{req2}
    \item Ability to test the models capability to disambiguate both \textbf{general purpose and domain-specific} senses; \label{req3}
    \item Ability to test the models capability to classify usages of previously \textbf{unseen words};
    \label{req4}
\end{enumerate}



Formally, each instance in the dataset consists of a target word $w$, a context $c$ containing the target word $w$, and its corresponding target sense $s$ represented by either its definition (Task 1), its hypernym/s (Task 2), or both definition and hypenyms (Task 3). The task aims to determine whether the intended sense of the word $w$ used in the context $c$ matches the target sense $s$. 

Table \ref{tab:examples} contains examples of instances from the WiC-TSV test set. 
Furthermore, a small sample of 10 
instances is available online in the form of a survey\footnote{\url{https://www.surveymonkey.com/r/LHYWXPV}}, where the achieved score is shown to the user after the submission.

\begin{table*}[t]
    \centering
    \small
    \scalebox{0.98}{
    \begin{tabular}{L{0.04\textwidth} L{0.40\textwidth} L{0.33\textwidth} L{0.14\textwidth} }
        \toprule
        \bf Tag & \bf Context & \bf Definition & \bf Hypernyms \\
        \toprule
        \multicolumn{4}{l}{\bf General Purpose (WNT/WKT)} \\
        \midrule
        T & \textbf{Smoking} is permitted . & the act of smoking tobacco or other substances	& breathing, 
        respiration, ventilation \\ 
        \midrule
        F & all that work went down the \textbf{sewer} & someone who sews & needleworker \\ 
        \bottomrule
        
        \toprule
        \multicolumn{4}{l}{\bf Cocktails (CTL)} \\
        \midrule
        T & We were 11 at table for this feast . We started the evening with \textbf{Bellini} , made with fresh , Niagara peaches . ( Thank you , Jack Lalanne Juicer ! ) & A Bellini cocktail is a mixture of Prosecco sparkling wine and peach pur\'ee. 
        & cocktail \\ 
        \midrule
        F & After a morning 's work I went off to see the \textbf{Bellini} retrospective at the Quirinale . Beautiful ! & A Bellini cocktail is a mixture of Prosecco sparkling wine and peach pur\'ee. 
        & cocktail \\ 
        \bottomrule
        
        \toprule
        \multicolumn{4}{l}{\bf Medical Subjects (MSH)} \\
        \midrule
        T & Italy now reports the second highest number of \textbf{corona} cases wordlwide . & A viral disorder characterized by high fever; cough; dyspnea; renal dysfunction and other symptoms of a viral pneumonia. 
        & 
        viral\_pneumonia; coronavirus\_ infection
        \\
        \midrule
        F & \textbf{Corona} Labs is happy to announce the general availability of the public beta of Android 64-bit Corona builds . & A viral disorder characterized by high fever; cough; dyspnea; renal dysfunction and other symptoms of a viral pneumonia. 
        & 
        viral\_pneumonia; coronavirus\_ infection 
        \\
        \bottomrule
        
        \toprule
        \multicolumn{4}{l}{\bf Computer Science (CPS)} \\
        \midrule
        T & pandas is a fast , powerful , flexible and easy to use open source data analysis and manipulation tool , built on top of the \textbf{Python} programming language . & Python is an interpreted, high-level, general-purpose programming language & object\_\allowbreak oriented\_\allowbreak programming\_\allowbreak language
       \\
        \midrule
          F & The present paper compares the recently studied \textbf{pythons} with those examined 20 years ago , and uses the combined dataset to assess the ecological sustainability . & Python is an interpreted, high-level, general-purpose programming language & object\_\allowbreak oriented\_\allowbreak programming\_\allowbreak language \\
        
        \bottomrule
    
    \end{tabular}
    }
    \caption{Sample instances from WiC-TSV. Target words are marked in bold. Tags: T (True) and F (False).}
    \label{tab:examples}
\end{table*}




\subsection{Dataset Construction}
\label{datasets}

In this section we detail the construction of the dataset. First, we describe the construction of the training and development set (Section \ref{train+dev const}) and then the test set (Section \ref{domain}), with a special focus on the creation of the domain-specific subsets.

\subsubsection{Training and Development Set}
\label{train+dev const}
Instances in the training and development set do not focus on a specific domain. 
As basis served the Word-in-Context (WiC) dataset \cite{pilehvar-camacho-collados-2019-wic}, which contains a target word $w$ and two contexts $c_1$ and $c_2$ for each instance. 
The contexts from WiC for noun instances come from two resources: WordNet and Wiktionary. 
To maintain the desirable characteristics of the WiC dataset (e.g., balanced data, not having repeated contexts across instances), the splits of the original training and development sets were treated separately in the following way: starting from a noun-only sub-sample, for each context $c_i$, the sense of the target word $w$ was mapped to the corresponding synset of WordNet, adding a sense identifier.
Each WiC instance was then split into two instances, one for each context.
For initial negative instances (i.e. $w$ has different intended senses in $c_1$ and $c_2$), the sense identifiers of these two instances were switched. To avoid information leakage, only one of the two instances were kept for the WiC-TSV dataset\footnote{If both instances were kept, the label could have been predicted with a high accuracy by counting the appearance of the target sense (even=True, odd=False).}. Finally, for each sense, the definition and hypernyms were derived from WordNet using the sense identifiers.\footnote{WordNet sense identifiers are omitted in the final dataset.}

\subsubsection{Test Sets}
\label{domain}

To make the dataset more challenging and realistic, the test set incorporates both general purpose and domain-specific instances.

\paragraph{General Purpose (WNT/WKT).}
The general purpose instances were generated analogously to \ref{train+dev const}. Hence, this test set is composed of both WordNet and Wiktionary examples, with definitions and hypernyms extracted from WordNet.

In the following we describe the construction of the domain-specific subsets.
The main difference between domain-specific and WNT/WKT test sets is that in the former the target sense remains the same. That means, that even though ``fork'' might have different senses within the computer science domain, we are only interested in one of these senses.

\paragraph{Cocktails (CTL).}
For the cocktails instances the target words were taken from the ``All about cocktails'' thesaurus\footnote{\url{vocabulary.semantic-web.at/cocktails}}. The thesaurus contains 300 entries describing not only cocktails, but also beverages, garnishes and glassware, among others. For instances obtained from this resource, the hypernym ``cocktail'' is used in the WiC-TSV dataset, while the definition is derived from the thesaurus.

\paragraph{Medical Subjects (MSH).}
For medical subject instances we use terms, definitions and hypernyms from the MeSH  thesaurus\footnote{\url{www.nlm.nih.gov/mesh/}}. This thesaurus is used for indexing medical articles and therefore contains a wide variety of terms in this domain. 
We considered various types, such as diseases, symptoms and body parts as target words.

\paragraph{Computer Science (CPS).}
Target words in the computer science domain were gathered manually, without a readily available thesaurus. The definitions were derived from the lead section of the corresponding Wikipedia page, while hypernyms were created by the consensus of two domain experts.
\paragraph{}

In order to create the domain-specific instances, first a list of ambiguous words and their domain-specific target senses was fixed for each domain. Then, we used the Wikilinks dataset \cite{singh12:wiki-links} as a basis for collecting different contexts containing the target words.
This dataset contains documents -- blog posts scraped from the web -- and the links from these documents to the Wikipedia pages, which were used to assign the intended sense (i.e., target sense or other sense) to the target word.
Where needed, additional contexts were collected manually by incorporating a search engine to find contexts for the target word. The intended senses for these instances were assigned manually.

\paragraph{Postprocessing.}
After creating the initial domain-specific instances, the subsets were checked manually to remove non-suitable and unsolvable instances. 
To maintain a rather realistic evaluation setup, data was not completely cleaned, meaning that contexts can contain \textit{noisy} elements such as headings or meta-info
derived from the websites (e.g., ``posted by'').

\subsection{Data Cleaning}
\label{data cleaning}
While the quality of the domain-specific instances is assured due to their manual creation process, an additional data cleaning step was introduced in which general purpose instances were manually curated. The instances from the test set were split into four sets with an overlap of 20\%. Each set was evaluated by an annotator regarding correctness and solvability of the instances. For example, when the hypernym of an instance was too generic to help in the disambiguation process, or the context itself was too ambiguous, the instance was marked as ``to filter out''. Each marked instance was reviewed by a second annotator, who could either confirm, or reject the request of removal. Instances marked by both annotators were removed. 

An example of such a removed instance would be the context \textit{``The \textbf{zero} sign in American Sign Language is considered rude in some cultures .''} for the target word ``zero'' with the target definition '\textit{a mathematical element that when added to another number yields the same number}'. 
In American Sign Language (ASL), ``zero sign'' is a ring-shaped hand sign using the thumb and pointing finger, similar to the OK-gesture. 
The provided instance mixes two senses of ``zero sign''. On the one hand, it refers to the hand gesture itself (synonymous to OK-gesture) which does not fully match the target sense. On the other hand, it also refers to the sign of the digit zero in ASL, which does match the target sense.

Other examples of filtered instances involve sentences where the target word may have been used metaphorically.

This procedure resulted in 106 instances which were removed. About 8\% of these instances were part of evaluation sets created to measure the human performance (see \ref{human-performance})\footnote{Annotations for these instances were removed before calculating the metrics presented in \ref{human-performance}}: 
the annotators achieved a mean accuracy of only 56\% on these instances.
This shows that the data cleaning step was necessary in order to ensure the data quality of the test set.

\begin{table}[t]
\setlength{\tabcolsep}{7.0pt}
\scalebox{0.85}{ 
\centering
\begin{tabular}{l|L{0.42\linewidth}|ccc}
\toprule
\bf   &
\bf &
\bf Total &
\bf $N_w$ &
\bf $R_+$ \\
\midrule
   \textbf{Train} & \textbf{WNT/WKT} & 2137 & 864 & 0.56 \\ 
      \midrule
  \textbf{Dev} & \textbf{WNT/WKT} & 389 & 377 & 0.51 \\
     \midrule
   \multirow{6}{*}{\textbf{Test}} & \textbf{General-domain} (\textbf{WNT/WKT}) & 717 & 664 & 0.54 \\
   \cmidrule(lr){2-5} 
     &\textbf{Domain-specific (MSH+CTL+CPS)}  & 589 & 25 & 0.47 \\
    \cmidrule(lr){2-5}
  &\textbf{MSH}  & 205 & 8 & 0.52 \\
  &\textbf{CTL}  & 216 & 9 & 0.43 \\
  &\textbf{CPS}  & 168 & 8 & 0.46\\

    \cmidrule(lr){2-5} \morecmidrules  \cmidrule(lr){2-5}
  & \textbf{All}  & 1306 & 689 & 0.51 \\
\bottomrule
\end{tabular}
}
\caption{\label{tab:statsdataset} Statistics of training, development and testing splits of WiC-TSV, including total number of instances (Total), unique number of target words ($N_w$) and percentage of positive instances ($R_+$).}
\end{table}

\subsection{Statistics}
\label{statistics}
A statistical overview of the dataset and their splits is shown in Table \ref{tab:statsdataset}. The totality of 3832 available instances were split into train, development and test sets with a ratio of 56:10:34 which allows a sophisticated analysis of the generalisation capabilities of tested systems, while still providing an appropriately sized training set.
 
The test set contains around 55\% general purpose instances and 45\% from specific domains. For each domain, the number of unique target words is relatively low compared to the general domain subset, which results in a higher number of instances per target word. However, for domain specific words, a great variety of senses is used in the contexts, yielding a big diversity among the instances. 
For all three splits, positive and negative instances are approximately balanced.


\subsection{Human Performance}
\label{human-performance}
To estimate the human performance upper bound, a sub-sample of the test set was manually annotated. The performance was evaluated on the setting of Task 3, meaning that both the definition and the hypernyms were provided to disambiguate. A random selection of 250 instances were split into two evaluation sets of the size of 150, resulting in a 20\% overlap. Each evaluation set was assigned to a non-expert annotator with English as native language.
No additional information - especially not from the respective ontology or about other senses of the target - was provided to the annotators and they were instructed not to use external knowledge sources (e.g. if they are not familiar with the domain-specific sense of a word). 

Results of the human performance evaluation can be found in Table \ref{tab:human_performance}.
The mean accuracy for the evaluated datasets was 85\%, with individual scores of 81\% and 89\%. To estimate the inter-annotator reliability, the agreement of the two annotators on the overlapping instances was calculated: for 42 instances (84\%) the annotators agreed on the label.

When evaluating the instances per domain, it can be seen that the general purpose instances were more difficult than the domain-specific ones, as annotators achieved an average accuracy of 82\% (individual scores of 77\% and 87\%) on the general purpose instances, while the mean accuracy on the domains were 89\% (83\% and 96\%), 92\% (88\% and 96\%), and 86.5\% (89\% and 84\%) for {MSH}, {CTL}, and {CPS}, respectively.
This performance difference is even more evident when comparing to the performance of non-native speakers: an additional experiment showed, that evaluators whose mother language is not English only achieved an average accuracy of about 77\% on the {WNT/WKT} instances, while performances on the domain specific subsets were comparable to native speakers.

\begin{table}
\centering
\setlength{\tabcolsep}{12.0pt}
\scalebox{0.96}{ 
\begin{tabular}{l|c}
\toprule

\bf   &
\bf Human Perf. 
\\
\midrule
  \textbf{WNT/WKT}	 & 82.1 	\\
   \textbf{MSH}	  		 	&  89.1\\
   \textbf{CTL} 				  &  92.0\\
  \textbf{CPS}   			&  86.5\\
   \midrule
   All 			&  85.3 \\
  
\bottomrule
\end{tabular}
}
\caption{\label{tab:human_performance} Average human accuracy for native English annotators, on different subsets of the dataset: general purpose, i.e., {WNT/WKT}, and the domain specific, i.e., {MSH}, {CTL}, and {CPS}.}
\end{table}

\section{Experimental Results}
\label{experiments}

In this section we evaluate the performance of different baseline models on our WiC-TSV benchmark. For our experiments we considered two main 
systems
, namely BERT \cite{devlin-etal-2019-bert} and FastText \cite{joulin2017bag}, as well as unsupervised baselines adapted to the corresponding tasks in WiC-TSV. 

\subsection{Evaluation Tasks}
\label{tasks}

The benchmark provides three different tasks depending on the input information available: definition-based (Section \ref{definition}), hypernym-based (Section \ref{hypernym}), and both (Section \ref{task3}).

\subsubsection{Task 1: Definition Information}
\label{definition}
In this task, the goal is to identify if the intended sense of the target word in the {\it context} matches the target sense described by the {\it definition}.

In other words, the model has to check if the sense represented by the {\it definition} can fit within the given {\it context}.
For this task, the system is provided with a {\it context} (in which the target word is marked) along with a {\it definition} (which describes one of the possible senses of the word).

\paragraph{Baselines.}
The first baseline is based on the pre-trained transformer-based language model \textbf{BERT}\footnote{We used the implementation of BERT available at \url{https://github.com/CyberZHG/keras-bert} for the {\it base} (BERT-B) and {\it large} (BERT-L) pre-trained models.}. It consists of a simple classification layer on top of the BERT model which is responsible for encoding the input.
For this task, we concatenate the context and the definition and feed the whole sequence to BERT. Then, the classification layer takes as input the concatenation of three different vectors, all provided by BERT: the \textsc{[CLS]} token representation, the representation of the target word in the context and the average representation of the words in the definition. This is similar to the baseline BERT model employed in SuperGLUE \cite{wang2019superglue}. It is worth mentioning that BERT is originally trained using WordPiece tokenization \cite{wu2016google}, which means that each word can be broken down into more than one sub-word. Therefore, in order to have a fixed length representation for each word, we take the average of its sub-word representations. Finally, the whole model is fine-tuned on the training set.

For the \textbf{FastText}-based baseline, we first extract the corresponding embeddings for each word in the context and definition, respectively. Then, the representation is simply computed as the average of the corresponding embeddings it contains. Next, these two representations are concatenated together to form a fixed length vector which we then feed to a fully connected layer. Finally, we put a simple classification layer on top of this fully connected layer and train the model on the training set.

We also evaluated \textbf{GlossBERT} \cite{huang-etal-2019-glossbert} on our dataset.
The authors describe a weak supervision algorithm that consists in surrounding the target word with special symbols -- quotation marks are used in the available implementation\footnote{\url{https://github.com/HSLCY/GlossBERT}}. We provide results both with (GBERT$_{ws}$) and without (GBERT) weak supervision. We chose the hyper-parameters as suggested by the authors, trained for 6 epochs and achieved the highest scores on the 4th epoch.

Unsupervised baselines \textbf{U-BERT} and \textbf{U-dBERT}, which do not make use of the training set, are simple threshold-based classifier which take the cosine distance of a target word representation and a definition representation into account. As source for these vectors serve BERT and DistilBERT, respectively. 

Similar to before, we derive the target word vector by taking the embedding of the target word in the context and the definition vector by averaging over all embeddings of the definition. The threshold is tuned on the development set with a step size of 0.02.

\subsubsection{Task 2: Hypernym Information}
\label{hypernym}

For this task, the system is provided with a target word (in a {\it context}) and a set of {\it hypernyms} for the target sense.
The task is to identify if the intended sense given through the context is the hyponym of the provided hypernyms. 
Note that, unlike Task 1, no {\it definition} is involved in this setting and the task is directed only by hypernymy information.

\paragraph{Baselines.}
We used baseline models similar to those used in the previous task. The only difference lies in how we shape the inputs fed to these models.

For the supervised and unsupervised BERT-based models, we put together the context with the hypernyms to form the input.
Similarly, for the FastText-based model, the hypernyms' embeddings are concatenated with the context's representation and fed to the classifier.

\subsubsection{Task 3: Both Sources of Information}
\label{task3}

In the third task systems are provided with both definition and hypernymy information.

\paragraph{Baselines.}
For this task, we concatenate the definition and the hypernyms, and feed the generated sequence together with the context to BERT. Then, the concatenation of the [CLS] token representation, the representation of the target word in the context and the average representation of the words in the definition/hypernyms sequence is fed to the classification layer.

For the unsupervised model we use the same BERT input and take the representation of the target in context and the average over the definition and hypernyms as input vectors.
For the FastText-based baseline, the hypernyms' embeddings are concatenated with both the context's representation and the definition representation and the combination is fed to the classifier.

\begin{table}
\setlength{\tabcolsep}{9.0pt}
\scalebox{0.8}
{ 
{
\noindent
\begin{tabular}{ll cc}
\toprule
\multirow{2}{*}{\bf } &     &
\multicolumn{2}{c}{\bf WiC-TSV}  \\

\cmidrule(lr){3-4}

\bf     &   & 
\bf  Acc  & \bf Prec~~~~~ Rec~~~~~~~F1 \\
\midrule

\multirow{7}{*}{\shortstack[l]{Task-1 \\ (def)}} &
BERT-B   &   75.3   &   71.7~~~~~ 84.9~~~~~77.7\\
& BERT-L &  75.3   &   70.4~~~~~ \textbf{88.5}~~~~~78.4\\
& FastText   &   53.7   &   54.1~~~~~ 57.6~~~~~55.7\\
& GBERT   &   \textbf{76.0}   &   71.3~~~~~ 88.2~~~~~\textbf{78.8}\\
& GBERT$_{ws}$   &   75.9   &   71.2~~~~~ 88.1~~~~~\textbf{78.8}\\
\cmidrule(lr){2-4}
& U-dBERT   &   56.9   &   \textbf{76.0}~~~~~ 22.0~~~~~34.2\\
& U-BERT   &   54.4   &   73.1~~~~~ 16.0~~~~~26.2\\

\midrule

\multirow{5}{*}{\shortstack[l]{Task-2 \\ (hyp)}} &
BERT-B   &   71.4   &   67.7~~~~~ 83.5~~~~~74.8	\\
& BERT-L &  \textbf{75.3}   &   \textbf{71.7}~~~~~ \textbf{85.1}~~~~~\textbf{77.8}\\
& FastText   &   52.7   &   52.4~~~~~ 73.6~~~~~61.1	\\
\cmidrule(lr){2-4}
& U-dBERT   &   62.3   &   64.8~~~~~ 56.3~~~~~60.2\\
& U-BERT   &   62.8   &   65.9~~~~~ 55.2~~~~~60.1\\

\midrule

\multirow{5}{*}{\shortstack[l]{Task-3 \\ (both)}} &
BERT-B   &   \textbf{76.6}   &   \textbf{74.1}~~~~~ 82.8~~~~~78.2	\\
& BERT-L &  76.3   &   72.6~~~~~ \textbf{85.7}~~~~~\textbf{78.6}\\
& FastText   &   53.4   &   52.8~~~~~ 79.4~~~~~63.4	\\
\cmidrule(lr){2-4}
& U-dBERT   &   61.2   &   70.6~~~~~ 40.3~~~~~51.3\\
& U-BERT   &   60.5   &   68.0~~~~~ 41.9~~~~~51.9\\

\midrule

\multicolumn{2}{l}{Baseline\textsubscript{True}}   &   50.8   &   50.8~~~~~~100~~~~~~67.3	\\
\it Human   & &  \textit{85.3}   &   \textit{80.2}~~~~~ \textit{96.2}~~~~~~\textit{87.4}	\\

\bottomrule
\end{tabular}
}
}
\caption{\label{tab:results_task_all}
Test set performance 
of the baseline models on WiC-TSV, in terms of accuracy, precision, recall, and F1, on the three different tasks.
Baseline\textsubscript{True} is a naive baseline that always returns \textit{True} and the human performance is computed as described in Section \ref{human-performance}.
} 
\end{table}

\subsection{Results}
\label{results}


\begin{table*}
\setlength{\tabcolsep}{4.2pt}
\scalebox{0.75}
{ 
{
\begin{tabular}{ll cc c cc c cc c cc}
\toprule
\multirow{2}{*}{\bf } &     &
\multicolumn{2}{c}{\bf WNT/WKT}    & &  
\multicolumn{2}{c}{\bf CTL} & &
\multicolumn{2}{c}{\bf MSH} & &
\multicolumn{2}{c}{\bf CPS} \\

\cmidrule(lr){3-4}
\cmidrule(lr){6-7}
\cmidrule(lr){9-10}
\cmidrule(lr){12-13}

\bf     &   & 
\bf  Acc  & \bf P ~~~~~~~ R ~~~~~~~ F1  & &
\bf  Acc  & \bf P ~~~~~~~ R ~~~~~~~ F1  & & 
\bf  Acc  & \bf P ~~~~~~~ R ~~~~~~~ F1  & &
\bf  Acc  & \bf P ~~~~~~~ R ~~~~~~~ F1  \\
\midrule

\multirow{7}{*}{\shortstack[l]{T1\\def}} &
BERT-B   &   73.3   &   74.0~~~~~ 77.7~~~~~75.8  &      &   \textbf{76.2}   &   {65.1}~~~~~ \textbf{98.9}~~~~~\textbf{78.4}   &      &   77.6   &   73.4~~~~~ 89.0~~~~~80.4   &      &   {80.0}   &   {70.5}~~~~~ {97.9}~~~~~{81.9} \\
& BERT-L   &   \textbf{77.1}   &   \textbf{75.7}~~~~~ \textbf{84.7}~~~~~\textbf{80.0}  &      &   73.1   &   62.3~~~~~ 95.3~~~~~75.4   &      &   75.3   &   70.6~~~~~ 89.6~~~~~78.9   &      &   70.2   &   61.4~~~~~ 97.4~~~~~75.3 \\
& FastText   &   56.2   &   58.9~~~~~ 61.9~~~~~60.3   &      &   49.8   &   39.0~~~~~ 30.8~~~~~34.3   &      &   51.7   &   52.2~~~~~ 79.2~~~~~62.9   &      &   50.4   &   45.6~~~~~ 38.5~~~~~41.6 \\
& GBERT   &   75.7   &   74.9~~~~~ 82.6~~~~~78.6   &      &   75.5   &   64.9~~~~~ 93.5~~~~~76.7   &      &   74.1   &   67.5~~~~~ 96.2~~~~~79.4   &      &   79.8   &   70.0~~~~~ \textbf{98.7}~~~~~81.9 \\
& GBERT$_{ws}$   &   75.2   &   74.6~~~~~ 81.6~~~~~78.0   &      &   70.4   &   59.7~~~~~ 95.7~~~~~73.6   &      &   \textbf{78.5}   &   71.5~~~~~ \textbf{97.2}~~~~~\textbf{82.4}   &      &   \textbf{82.7}   &   73.3~~~~~ \textbf{98.7}~~~~~\textbf{84.2} \\ \cline{2-13}
& U-BERT   &   49.2   &   64.1~~~~~ 13.0~~~~~21.6   &      &   57.4   &   \textbf{100}~~~~~~~~ 1.1~~~~~~~2.1   &      &   62.0   &   78.0~~~~~ 36.8~~~~~50.0   &      &   63.1   &   \textbf{100}~~~~~ 20.5~~~~~34.0 \\
& U-dBERT   &   51.5   &   67.0~~~~~ 19.4~~~~~30.1   &      &   56.9   &   0.0~~~~~~~~ 0.0~~~~~~~0.0   &      &   65.9   &   \textbf{86.0}~~~~~ 40.6~~~~~55.1   &      &   69.0   &   93.3~~~~~ 35.9~~~~~51.9 \\

\midrule
\multirow{5}{*}{\shortstack[l]{T2\\hyp}} &
BERT-B   &   68.6   &   70.0~~~~~ 72.9~~~~~71.4   &      &   77.9   &   66.6~~~~~ {97.8}~~~~~79.3   &      &   71.9   &   65.1~~~~~ \textbf{98.4}~~~~~78.3   &      &   74.4   &   64.7~~~~~ \textbf{98.7}~~~~~78.2    \\
& BERT-L   &   \textbf{71.4}   &   \textbf{71.7}~~~~~ \textbf{77.4}~~~~~\textbf{74.4}  &      &   \textbf{82.7}   &   \textbf{72.7}~~~~~ 96.4~~~~~\textbf{82.8}   &      &   77.2   &   71.2~~~~~ 94.0~~~~~\textbf{81.0}   &      &   \textbf{80.6}   &   \textbf{71.3}~~~~~ 97.4~~~~~\textbf{82.3}   \\
& FastText   &   56.8   &   58.9~~~~~ 66.3~~~~~62.1   &      &   43.1   &   43.0~~~~~ \textbf{99.3}~~~~~60.0   &      &   49.1   &   50.4~~~~~ 84.0~~~~~62.9   &      &   52.0   &   48.8~~~~~ 65.0~~~~~55.3    \\ \cline{2-13}
& U-BERT   &   57.6   &   61.3~~~~~ 57.8~~~~~59.5   &      &   57.4   &   52.0~~~~~ 14.0~~~~~22.0   &      &   74.6   &   79.3~~~~~ 68.9~~~~~73.7   &      &   77.4   &   77.0~~~~~ 73.1~~~~~75.0 \\
& U-dBERT   &   55.0   &   58.1~~~~~ 58.5~~~~~58.3   &      &   56.0   &   42.9~~~~~~ 6.5~~~~~~11.2   &      &   \textbf{80.5}   &   \textbf{86.7}~~~~~ 73.6~~~~~79.6   &      &  79.2   &   75.9~~~~~ 80.8~~~~~78.3 \\

\midrule
\multirow{5}{*}{\shortstack[l]{T3\\both}} &
BERT-B   &   73.5   &   76.1~~~~~ 74.2~~~~~75.1   &      &   \textbf{79.2}   &   {67.8}~~~~~ {98.2}~~~~~\textbf{80.2}   &      &   \textbf{79.8}   &   75.8~~~~~ {89.6}~~~~~\textbf{82.1}   &      &   \textbf{82.1}   &   {73.0}~~~~~ \textbf{97.9}~~~~~\textbf{83.6}  \\
& BERT-L   &   \textbf{77.3}   &   \textbf{77.2}~~~~~ \textbf{82.1}~~~~~\textbf{79.6} &      &   76.4   &   67.0~~~~~ 90.0~~~~~76.6   &      &   75.4   &   71.6~~~~~ 87.4~~~~~78.7   &      &   72.8   &   63.8~~~~~ 96.2~~~~~76.7   \\
& FastText   &   57.1  &   58.0~~~~~ 74.0~~~~~65.0   &      &   43.1   &   43.1~~~~~ \textbf{100}~~~~~60.2  &      &   51.1   &   51.5~~~~~ \textbf{90.3}~~~~~65.6   &      &   54.0   &   50.5~~~~~ 67.1~~~~~57.3   \\ \cline{2-13}

& U-BERT   &   54.4   &   61.3~~~~~ 41.5~~~~~49.5   &      &   58.8   &   62.5~~~~~ 10.8~~~~~18.3   &      &   71.2   &   78.3~~~~~ 61.3~~~~~68.8   &      &   75.6   &   \textbf{87.8}~~~~~ 55.1~~~~~67.7 \\
& U-dBERT   &   54.8   &   62.4~~~~~ 40.4~~~~~49.1   &      &   57.9   &   \textbf{75.0}~~~~~~~ 3.2~~~~~~~6.2   &      &   74.1   &   \textbf{87.3}~~~~~ 58.5~~~~~70.1   &      &   76.8   &   86.8~~~~~ 59.0~~~~~70.2 \\

\midrule

\multicolumn{2}{l}{Baseline\textsubscript{True}}  & 53.8   &   53.8~~~~~ 100~~~~~70.0   &      &   43.1   &   43.1~~~~~ 100~~~~~60.2   &      &   51.7   &   51.7~~~~~ 100~~~~~68.2   &      &   46.4   &   46.4~~~~~ 100~~~~~63.4  \\

\bottomrule
\end{tabular}
}
}
\caption{\label{tab:results_task_domain}
Performance 
for the baseline models for the three tasks (i.e., T1: definition-based, T2: hypernymy-based, and T3: both sources of information) split by domain: General ({WNT/WKT}), Cocktails ({CTL}), Medical Subjects ({MSH}), and Computer Science ({CPS}). Baseline\textsubscript{True} is a naive baseline that always returns ``True''. Human performance in terms of accuracy is estimated to be 82.1\% ({WNT/WKT}), 92.0\% ({CTL}), 89.1\% ({MSH}) and 86.5\% ({CPS}) as described in Section \ref{human-performance}.}
\end{table*}

Table \ref{tab:results_task_all} shows the overall results for the three tasks. As can be observed, GlossBERT performs best in terms of accuracy and F$_1$. BERT-L is a little worse, but achieves the best recall. 
The worst supervised baseline -- FastText -- does not perform better than a naive baseline that retrieves all instances as true. This also reinforces the challenging nature of the benchmark, as even BERT-based models are far from the human annotator performance (estimated on 85.3\% for accuracy). Clearly, the definition information is more helpful than the hypernyms for BERT, while the combination of both attains the best overall results. Yet GlossBERT reaches a better performance with definition only\footnote{We did not evaluate GlossBERT with hypernyms as suchconfiguration was not considered by the authors in the original system and its integration would not be straightforward.}. 

The unsupervised models only perform well with hypernyms. Though U-dBERT reaches the best precision in Task 1, the recall remains very low and therefore the overall performance.

Another point to highlight is the high recall of BERT-based models, in contrast to its precision.
This is mainly attributed to the domain-specific subsets as it will be analysed below. As for the comparison between BERT-based models, the larger model (BERT-L) performs as expected better than the base model (BERT-B) overall.

\subsection{Analysis}

In order to better understand the results, in this section we perform a focused analysis on the performance split by domain.

\subsubsection{Domain-based Analysis}

Table \ref{tab:results_task_domain} presents the results split by domain. FastText faces a massive challenge in adapting to new domains and generalising from WNT/WKT to the other domains. However, BERT-based models show to be much more robust to domain changes. In fact, the results on the domain-specific instances are in the same ballpark as the {WNT/WKT} test set.  
This can be attributed to the fact that specific domains highly constrain the set of possible senses for a word, resulting in an easier WSD classification task \cite{magnini2002role}. 
On the other hand, WordNet is known to be quite fine-grained (e.g., the noun \textit{run} has 16 different senses in WordNet).

Surprisingly, unsupervised DistilBERT achieves the best accuracy over all tasks and classifiers on MSH. However, both unsupervised models do not perform well on WNT/WKT and CTL. We can observe that supervised models are significantly more reliable and produce similar scores on different tasks and datasets 
than unsupervised models.

In general, for BERT-based models, recall is substantially higher than precision on the domain-specific subsets. This is desirable in a retrieval setting where a high-coverage retrieval of relevant cases is of more importance.
Interestingly, among the two BERT alternatives, the smaller model performs better on the domain-specific subsets, suggesting that it is more robust to domain changes.
This is an important observation which needs further careful investigation in future work, given that most evaluation benchmarks (on which the larger model consistently outperforms the smaller one) comprise in-domain test sets, which cannot reveal robustness across domains.

\subsubsection{In-domain Few-shot Analysis} 

Although the availability of big annotated domain-specific training sets is quite rare, the presence of a small training set forms a realistic scenario.
Incorporating these domain-specific instances in the model training could potentially increase its prediction performance. To investigate this theory, we performed an additional analysis focusing on the usage of in-domain examples in the learning process, where for each domain 100 instances from the test sets were used as a training set. To enforce the assumption that not all target senses would be seen during the training process, we put aside all instances of 3 target words for each domain test set.\footnote{To add robustness to the results, three different random samples were considered for this experiment, with the results being averaged after the three different runs.}
Two additional domain-based strategies were considered: (1) \textit{few-shot learning:} only using the domain-specific instances, and (2) \textit{continued learning:} extending the existing general-purpose training set with the domain-specific instances. 

For this analysis we focused on Task-3 and BERT-large, which performed better overall. Table \ref{tab:results_in_domain_f1} shows the F1 results. In general, 
few-shot learning
works surprisingly well overall (achieving the best overall performance in the CTL and MSH domains). On CTL pairs unseen during training, it even performs considerably better  than the same BERT model trained in the continued learning setting.
In the CPS domain, for both few-shot in-domain learning and continued learning the performance on seen target words is quite high, while the prediction of unseen target words produces relatively low F1 scores, which indicates a low ability to generalise to new senses.
As for the model trained on the general-domain dataset, it performs best in the CPS domain, but performs considerably lower than the domain-tuned alternatives in the CTL domain. Indeed, the domain-tuned BERT systems clearly outperform the same model trained on the general domain on seen pairs, proving the importance of obtaining word-specific examples to boost performance. 
However, this may not be realistic in practice, and therefore further research should be devoted in improving the generalization capabilities of disambiguation systems, and language models in particular.
These findings are consistent with the results of an experiment conducted with GlossBERT in the few-shot learning setting on Task-1: the overall accuracy increase ranged from $0.1$\% (CPS) to $13.6$\% (CTL) compared to the model trained solely on general domain instances.

\begin{table}[]
\setlength{\tabcolsep}{3.3pt}
\resizebox{\columnwidth}{!}
{
\begin{tabular}{llllllllll}
\toprule
\multicolumn{1}{c}{\multirow{2}{*}{\textbf{Train}}} & \multicolumn{3}{c}{\textbf{CTL}}        & \multicolumn{3}{c}{\textbf{MSH}}           & \multicolumn{3}{c}{\textbf{CPS}}  \\ \cmidrule(lr){2-4} \cmidrule(lr){5-7} \cmidrule(lr){8-10}
\multicolumn{1}{c}{}                                   & \textbf{All} & \textbf{See} & \textbf{Uns} & \textbf{All} & \textbf{See} & \textbf{Uns} & \textbf{All} & \textbf{See} & \textbf{Uns} \\ \midrule
\textbf{WNT}                                        & 76.5         & 74.7          & 75.5            & 77.4         & 82.9          & 76.0              & \textbf{74.9}         & 77.8          & \textbf{74.8}            \\ 
\textbf{WNT+D}                       & 80.2         & \textbf{89.5}          & 75.5            & 78.5         & 86.2          & \textbf{76.6}            & 73.6         & 88.9          & 72.9            \\
\textbf{Dom}                            & \textbf{84.2}         & 88.8          & \textbf{82.6}            & \textbf{78.7}         & \textbf{88.6}          & 75.5            & 70.7         & \textbf{93.3}          & 69.1            \\ 
\midrule
\textbf{Base\textsubscript{True}}                                   & 58.5         & 54.8          & 59.1            & 68.2         & 70.9          & 67.5            & 62.6         & 58.3          & 62.8            \\ 
\bottomrule
\end{tabular}
}
\caption{\label{tab:results_in_domain_f1}
F1 score for the in-domain few-shot analysis (Task-3) using BERT-L trained on general domain (WNT), domain-specific (Dom) and general domain fine-tuned on the target domain (WNT+D). 
In addition to 
the full test set (All), 
results are split on seen (See) and unseen (Uns), as per the presence or absence of the target word in the domain-specific training set.
}
\end{table}

\section{Conclusions and Future Work}
\label{conclusion}

In this paper we have introduced the Target Sense Verification task, a re-formulation of WSD where the equivalence of the intended sense of a word in context and a single given sense is evaluated. Furthermore, we presented WiC-TSV, a multi-domain benchmark which differs from existing WSD datasets in three main ways: (1) it is based on TSV and therefore framed as a binary classification task where only one target sense needs to be verified, (2) it is independent from external sense inventories, and (3) its test set contains instances from three specific and heterogeneous domains are included: cocktails, medical subjects and computer science. 
Our benchmark therefore opens the floor for different disambiguation algorithms that do not require modeling the entirety of a sense inventory.
This characteristic also provides a crucial advantage in enterprise and domain-specific settings
as it facilitates the development of systems which are only aimed at modelling the domain at hand. 
Moreover, having these out-of-domain test instances makes our benchmark more robust and generalisable, preventing (or making it harder) for statistical models to learn spurious correlations from the training set, which has been proven to be an issue in standard NLP tasks \cite{poliak-etal-2018-hypothesis,gururangan-etal-2018-annotation,linzen2020can}.

In our initial experiments we found that current state-of-the-art disambiguation techniques based on pre-trained language models such as BERT are very accurate at handling ambiguity, even in specialised domains. However, there is still room for improvement as highlighted by the gap with the human performance. 
This benchmark therefore opens up avenues for future research on domain-transfer and on developing general-purpose solutions which can perform well on a variety of domains without the need for large amounts of training data. 

As future work, we are planning to further investigate and analyse the robustness of pre-trained models with respect to domain changes.
Also, it would be interesting to develop hybrid models which take both definition and hypernymy information into account -- in this paper we combined both sources in BERT in a simple manner, but more complex models should lead to further improvements.




\bibliography{anthology,eacl2021}
\bibliographystyle{acl_natbib}

\end{document}